\documentclass[letterpaper, 10 pt, conference]{ieeeconf}
\IEEEoverridecommandlockouts
\usepackage{amsmath}
\usepackage{amssymb}
\usepackage{balance}
\usepackage{booktabs}
\usepackage{graphicx}
\usepackage{siunitx}
\usepackage{textcomp}
\usepackage[absolute, showboxes]{textpos}
\graphicspath{{./img/}}
\DeclareSIUnit\vehicle{veh}
\DeclareSIUnit\lane{lane}
\TPMargin*{3pt}

\newcommand\copyrighttext{
	\footnotesize
	\noindent
	\textcopyright\,2022 IEEE.
	Personal use of this material is permitted.
	Permission from IEEE must be obtained for all other uses, in any current or future media, including reprinting/republishing this material for advertising or promotional purposes, creating new collective works, for resale or redistribution to servers or lists, or reuse of any copyrighted component of this work in other works.}%
\newcommand\copyrightnotice{%
	\begin{textblock*}{7in}(0.75in,0.25in)
		\copyrighttext
	\end{textblock*}
}

\title{\LARGE \bf
Cooperative Behavior Planning for\\Automated Driving Using Graph Neural Networks
}

\author{
	Marvin~Klimke$^{1,2}$,
	Benjamin~V\"olz$^{1}$, and
	Michael~Buchholz$^{2}$%
\thanks{$^{1}$The authors are with the Robert Bosch GmbH, Corporate Research, D-71272 Renningen, Germany. E-Mail: {\tt\small\{marvin.klimke, benjamin.voelz\}@de.bosch.com}}%
\thanks{$^{2}$The authors are with the Institute of Measurement, Control and Microtechnology, Ulm University, D-89081 Ulm, Germany. E-Mail: {\tt\small michael.buchholz@uni-ulm.de}}%
\thanks{Part of this work was financially supported by the Federal Ministry for Economic Affairs and Climate Action of Germany within the program "Highly and Fully Automated Driving in Demanding Driving Situations" (project LUKAS, grant numbers 19A20004A and 19A20004F).}%
}

\begin{document}

\maketitle
\copyrightnotice
\thispagestyle{empty}
\pagestyle{empty}

\begin{abstract}
Urban intersections are prone to delays and inefficiencies due to static precedence rules and occlusions limiting the view on prioritized traffic.
Existing approaches to improve traffic flow, widely known as automatic intersection management systems, are mostly based on non-learning reservation schemes or optimization algorithms.
Machine learning-based techniques show promising results in planning for a single ego vehicle.
This work proposes to leverage machine learning algorithms to optimize traffic flow at urban intersections by jointly planning for multiple vehicles.
Learning-based behavior planning poses several challenges, demanding for a suited input and output representation as well as large amounts of ground-truth data.
We address the former issue by using a flexible graph-based input representation accompanied by a graph neural network.
This allows to efficiently encode the scene and inherently provide individual outputs for all involved vehicles.
To learn a sensible policy, without relying on the imitation of expert demonstrations, the cooperative planning task is considered as a reinforcement learning problem.
We train and evaluate the proposed method in an open-source simulation environment for decision making in automated driving.
Compared to a first-in-first-out scheme and traffic governed by static priority rules, the learned planner shows a significant gain in flow rate, while reducing the number of induced stops.
In addition to synthetic simulations, the approach is also evaluated based on real-world traffic data taken from the publicly available inD dataset.
\end{abstract}

\section{Introduction}
\label{sec:intro}

Urban traffic regularly exhibits disturbances and inefficiencies caused by simple traffic management schemes faced with large volumes of vehicles.
Especially smaller intersections are typically handled by static priority rules, resulting in vehicles approaching from a minor road having to yield.
Moreover, occlusions through buildings or other objects are highly prevalent in urban areas, limiting the view for both human drivers and vehicle-bound sensory systems.

The increasing use of connected vehicles (CVs) and connected automated vehicles (CAVs) opens up new opportunities to increase the traffic efficiency.
Those vehicles can announce their presence and possibly share perception data with surrounding road users via a communication link.
Moreover, with edge computing resources becoming available in urban areas, it is viable to build and maintain a local environment model of, e.g., an intersection and its surroundings.
Such an edge server can distribute the environment model to connected vehicles in the operational area.
CAVs, for instance, can make use of the information by incorporating it into their planning algorithms.

\begin{figure}
	\centering
	\includegraphics[width=\linewidth]{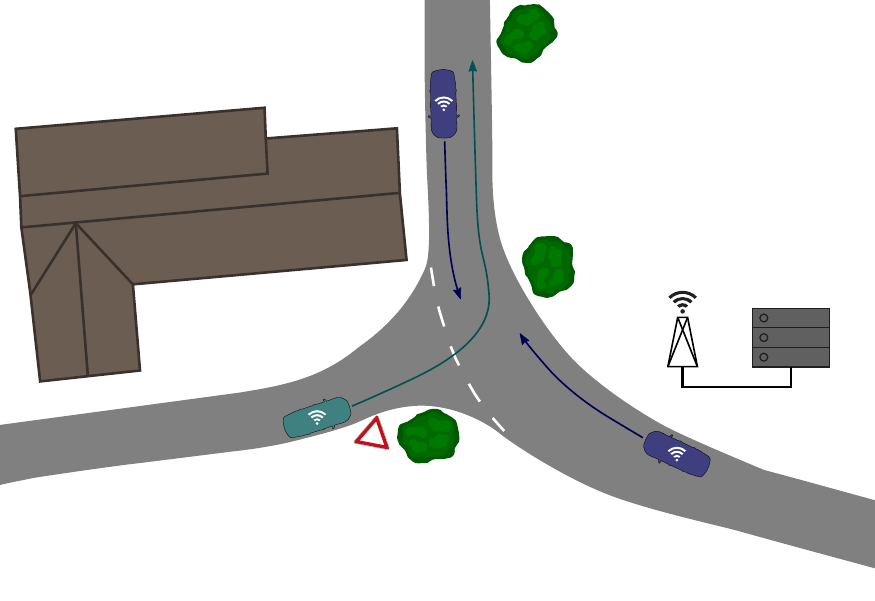}
	\caption{Cooperative maneuver at an urban intersection. The planning module on the edge server requests the blue vehicles on the main road to slow down. Hence, the turquoise vehicle can merge without having to stop.}
	\label{fig:intro}
\end{figure}

In the publicly funded project MEC-View, research on connected automated driving was conducted using a testing site at a suburban three-way intersection in the city of Ulm in Germany \cite{buchholz2021handling}.
Due to buildings occluding the view onto the priority road, an automated vehicle merging from the side road has to decelerate strongly, before being able to safely enter the intersection based solely on its own perception system.
With the support of the environment model provided by the edge server, the automated vehicle can transition smoothly onto the main road, given that appropriate space is available.
We build upon this approach and discuss the potential of multi-agent planning schemes that are executed on the server.
Based on the fused environment model, a cooperative plan for handling intersection traffic is derived that can be proposed to the connected vehicles as behavioral instructions.
Explicit deviations from static priority rules become possible.
For instance, vehicles on the main road can be requested to slow down and thus letting a vehicle from the side road merge into the emerging gap, as depicted in Fig.~\ref{fig:intro}.

Prior research on automatic intersection management (AIM) primarily focuses on non-learning algorithms, like reservation-based or optimization-based programs.
At the same time, machine learning-based approaches show remarkable results on prediction tasks in automated driving as well as planning for a single ego vehicle.
The lack of fitting ground-truth data prevents the application of supervised learning for cooperative behavior planning.
To bridge this gap, the present work proposes to train a reinforcement learning (RL) policy for multi-agent planning in a simulated environment, resulting in the following contributions:
\begin{itemize}
	\item Leveraging machine learning to perform cooperative multi-agent planning for urban automated driving,
	\item To the best of our knowledge, we propose the first AIM system exploiting graph neural networks,
	\item Evaluation based on real-world traffic data taken from a publicly available urban driving dataset.
\end{itemize}

The remainder of the paper is structured as follows:
Section~\ref{sec:sota} discusses related work in the field of AIM and machine learning-based planning for automated driving.
The proposed behavioral planning scheme is comprehensively introduced in Section~\ref{sec:approach}.
Afterwards, evaluation results obtained in synthetic simulations and based on real-world traffic data are given in Section~\ref{sec:eval}.
Section~\ref{sec:conclusion} concludes the paper and gives an outlook on future work.

\section{Related Work}
\label{sec:sota}

The analysis on the state of the art first considers existing approaches to AIM.
Because machine learning is seldom used for AIM, we subsequently investigate learning-based behavioral planning methods.
Due to the large body of existing literature, we present a selection of commonly used approaches and refer the reader to surveys for a more extensive overview.

\subsection{Automatic Intersection Management}
\label{ssec:aim}

Past research brought forth a variety of AIM schemes, surveyed for instance by \cite{zhong2020autonomous}.
The authors identify centralization as a crucial feature for distinguishing AIM schemes.
Thereby, a fully centralized scheme exhibits a single coordination unit that is in charge of planning the intersection traversal and acts as the communication partner for all vehicles.
In a fully distributed AIM, a cooperative plan is negotiated by the vehicles on their own.

A centralized, reservation-based AIM system is proposed in \cite{dresner2008multiagent}, which employs a first-in-first-out policy for assigning clearance to cross the intersection.
A driver agent places a request when its vehicle is about to enter the monitored intersection area, covering possible conflict points with other paths.
The intersection manager maintains tile-based reservations and confirms the request if the affected tiles are free.
This approach can be combined with a traffic light to enable the co-usage of the intersection by human drivers and automated vehicles.

Optimization-based intersection management systems are published, for instance, in \cite{malikopoulos2018decentralized} and \cite{kamal2015vehicle-intersection}.
Those works assume full penetration of CAVs that laterally follow predefined lanes on urban intersections.
The distributed energy-optimizing approach \cite{malikopoulos2018decentralized} further disallows turning maneuvers and the utilization of two conflicting paths at the same time.
In \cite{kamal2015vehicle-intersection}, the longitudinal control of vehicles is performed by a centralized intersection coordination unit employing a model predictive control (MPC) scheme.
Both works demonstrate efficiency gains in time and fuel consumption by comparison to a traditional signalized intersection.
A prevalent issue with optimization-based approaches is the unfavorable scaling of computational demand with increasing traffic density.
In \cite{mertens2022cooperative}, a novel AIM scheme is presented, that is also capable of handling mixed traffic, i.e., simultaneous usage by automated and human-driven vehicles.

The concept of platooning \cite{morales_medina2018cooperative} can also be used for AIM.
Based on a so-called virtual inter-vehicle distance, a pair of vehicles can adapt their velocities to cross their conflict point with sufficient clearance.
The authors acknowledge that significant adaptions would have to be made for managing an intersection under mixed traffic.

\subsection{Machine Learning-Based Planning}
\label{ssec:mlplan}

Machine learning-based approaches to automated driving experience rising interest of researchers during the last years.
A survey of recent deep reinforcement and imitation learning planning methods for a single ego vehicle can be found in \cite{zhu2021survey}.
The authors categorize published works by the type of input data (e.g. sensor measurements or object detections) and output representation (e.g. behavioral planning or direct control outputs).
Because individual sensor measurements are not suited for cooperative planning over multiple vehicles, we limit our analysis to methods that require a prior perception system to be in place.
Readers interested in machine learning-based prediction for automated driving are referred to comprehensive surveys on the topic, like \cite{lefevre2014survey}.
On the output side, multi-agent planning requires an intermediate representation that can be passed to various vehicles in the scene, laying the focus on high-level behavioral planning approaches.

Imitation learning describes the application of supervised learning techniques to automated driving by training on expert drivers' demonstrations, which can be obtained from datasets or accordingly equipped testing vehicles.
Based on object detections from a dedicated perception system and high-definition map information, a typical approach is to render the surroundings of the ego vehicle in a raster image that is subsequently processed by a convolutional neural network (CNN) \cite{chen2019deep,bansal2019chauffeurnet}.
To address the problem of distributional shift between training data and closed-loop test conditions, various improvements have been proposed, like perturbing a random subset of training trajectories to teach the model to recover from atypical states \cite{bansal2019chauffeurnet}.
Being based on supervised learning, those techniques share the large needs for high-quality training data.
This limits their prospective transfer to cooperative multi-agent planning because ground-truth data showing cooperative maneuvers is virtually not available.
Urban traffic datasets (e.g. the inD dataset \cite{bock2020ind}) show road users obeying to static priority rules or traffic lights; both entities that shall become obsolete with cooperative planning.

In contrast to supervised learning, RL approaches evade the requirement for large datasets by instead exploring possible actions in a simulated environment and exploiting a reward signal to learn the desired behavior.
In \cite{capasso2021end--end}, a driving policy for controlling the acceleration and steering angle is trained through RL that is applied to multiple vehicles in a common simulated environment.
As there is no explicit communication between the different vehicles' policies, no cooperation is shown in traffic.
Based on a raster image representation, this approach shares the unfavorable scaling of computational load with the number of participants in the scene, because each vehicle requires an individual image, centered on its pose for sensible inference.
An alternative RL-based approach to coordinated driving on an urban intersection was published in \cite{wu2019dcl-aim}.
By maintaining a tile-based reservation of the intersection, the decentralized policies can choose from the set of actions that do not cause a collision.
Apart from this limitation of the action space, there is no further inter-agent communication that could enable cooperative maneuvers.

When encoding the semantic environment of a vehicle in urban traffic, the number of potentially relevant entities (e.g. other vehicles) is highly dynamic.
This makes fixed-size network architectures and input representations often used in RL unsuitable for the task at hand.
In \cite{huegle2019dynamic}, it is proposed to encode input features per vehicle using a multilayer perceptron followed by a permutation invariant operation for pooling the resulting features.
The aggregated feature vector is then propagated through another fully connected network to finally infer actions for a single ego vehicle.
The authors extend their work to encode whole traffic scenes including lanes and traffic signs and compare it to using a graph convolutional network for the same task in \cite{huegle2020dynamic}.
Similarly, \cite{hart2020graph} proposes to encode the vehicles being present in the scene as graph vertices.
However, none of the described works can handle multi-agent planning.

\section{Proposed Approach}
\label{sec:approach}

In this section, our proposed approach is presented, beginning with a discussion on learning paradigms for multi-agent usage.
Afterwards, the graph-based input representation is introduced, followed by details on the network architecture and reward engineering.

\subsection{Learning Paradigm}
\label{ssec:learningparadigm}

In RL algorithms, an agent typically interacts with an environment in discrete time steps.
The agent observes the current state of the environment and subsequently chooses an action, whose effect is evaluated by a reward signal.
With multiple entities to be controlled, one can pursue different learning paradigms depending on the degree of centralization in multi-agent reinforcement learning \cite{gronauer2021multi-agent}.
Instead of having the various agents interact individually with the environment and learn independent policies, cooperative planning is modeled best by the centralized training centralized execution paradigm.
Because different agents, which shall take part in a cooperative maneuver, have to be able to communicate explicitly.
This holds not only during training, but also at inference time.
In paradigms relying on decentralized execution, the agents would have to learn an implicit communication scheme through their behavior.
With the fused environment model being available to the server-side planner, considering the planning problem over all vehicles in the scene using a joint RL agent allows for explicit communication and hence better cooperation.

Like many RL problems, the cooperative planning problem can be denoted as a Markov decision process (MDP), defined as the tuple $(S,\,A,\,T,\,R)$.
It consists of a set of states $S$ that fully describe the traffic scene at a given time. $A$ denotes the set of actions the RL agent can choose from while interacting with the environment.
The transition function $T(s\,,a\,,s')$ describes the probability of changing from state $s \in S$ to $s' \in S$ when applying action $a \in A$, whereas the reward signal is determined by the function $R(s,\,a)$.
Since the multi-agent planning problem contains different vehicles in the scene, the dimensionalities of the state space and the action space depend on the number of vehicles currently present and may vary over time.

\subsection{Input Representation}
\label{ssec:inputrepresentation}

\begin{figure}
	\centering
	\includegraphics[width=\linewidth]{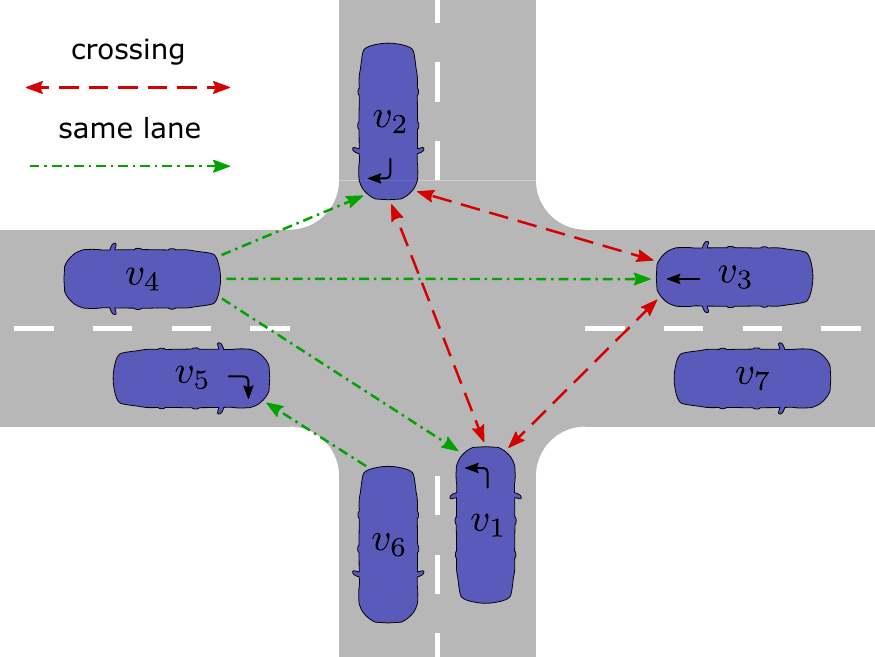}
	\caption{The graph-based input representation illustrated on an arbitrary traffic scene at a four-way intersection. The vehicles' turning intentions are denoted by arrows on their hood.}
	\label{fig:scene_graph}
\end{figure}

A well-suited input representation is crucial for applying artificial neural networks successfully.
We identify three major requirements for an input scene representation to be used in cooperative multi-agent planning:
\begin{itemize}
	\item Invariance on the number of vehicles in the scene,
	\item Permutation invariance of the input nodes,
	\item Permutation equivariance regarding the output nodes.
\end{itemize}
Simple tabular representations already lack the invariance properties.
The limitations of fixed-sized inputs for behavior planning are elaborated more extensively in~\cite{huegle2019dynamic}.
A rendered raster image of the scene, as often used for CNNs, fulfills the invariance requirements, but typically requires a target agent to be centered around \cite{chen2019deep}.
This process must be repeated to produce individual outputs for each agent, making the application to a large number of vehicles computationally infeasible.
Permutation equivariance means that the inferred outputs of given agents in the scene are independent of their ordering in the input vector.
Hence, our work proposes to use a lean and flexible graph-based scene representation, shown in Fig.~\ref{fig:scene_graph}, which fulfills all above requirements.
The current state of the environment is thus defined as $S=(V,\,E,\,U)$, with $V$ being the set of vertices corresponding to the vehicles in the scene and $E$ denoting edges depending on the pairwise relation between vehicles.
For each vehicle, one vertex in $V$ stores the corresponding input features.
Each of the directed edges is assigned one of two edge types in $U$, either \emph{same~lane}, or \emph{crossing}:
\begin{equation}
(v_1,\,r,\,v_2) \in E = V \times U \times V.
\end{equation}
Two vehicles in front of the intersection whose paths cross or merge are being connected bidirectionally with crossing edges, like $v_1$,~$v_2$, and~$v_3$ in the figure.
The same lane edge, in contrast, is used to connect two vertices of vehicles on the same path, pointing from the predecessor to the following one (e.g. $v_6$ and $v_5$).
This is motivated by the observation that vehicles should adapt their behavior to the preceding vehicle and not vice-versa.
Note that the graph does not need to be connected.
Some vehicles may form a disjoint sub-graph, if they share no conflicts with the remaining vehicles, as it is the case for $v_5$ and $v_6$ or $v_7$ in Fig.~\ref{fig:scene_graph}.

\begin{figure}
	\centering
	\includegraphics[width=\linewidth]{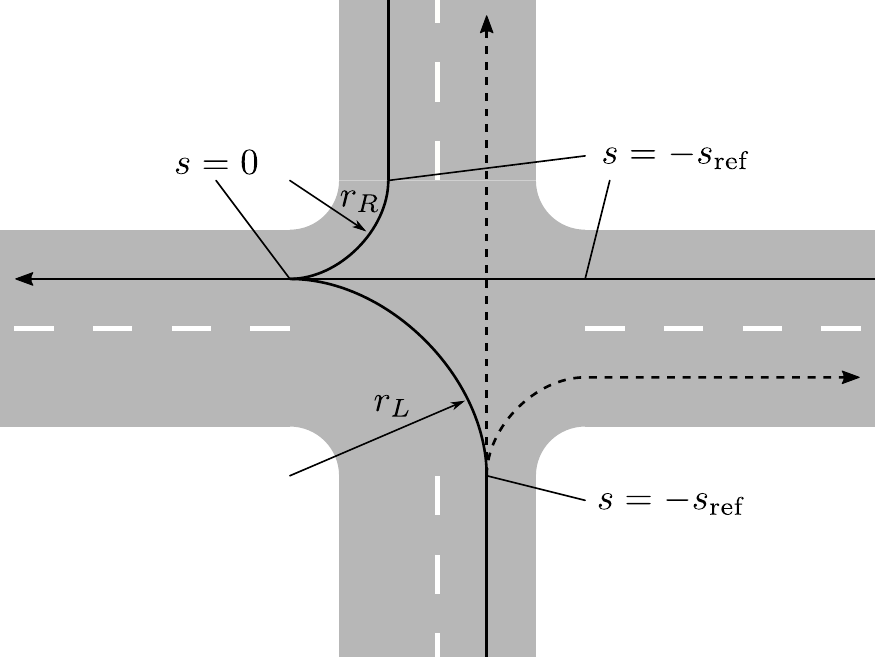}
	\caption{Path parameterization by a common reference length. Each path enters the intersection area at $s=-s_\mathrm{ref}$ and leaves it at $s=0$.}
	\label{fig:path_length}
\end{figure}

The input feature vector for each vehicle consists of three values and is denoted as $h^{(0)} = [s,\,v,\,d]^T$, where the upper index denotes the layer number.
The longitudinal position of the vehicle along its path is denoted by $s$, with $s=0$ defined as the point where the path leaves the intersection area.
This corresponds to the longitudinal coordinate of a Frenet coordinate pair.
Because different maneuvers (e.g. straight driving and right turns) cause a difference in the path length on the intersection, the effective path length is scaled to a common reference length $s_{\mathrm{ref}}$, as depicted in Fig.~\ref{fig:path_length}.
Thereby, the entry point to the intersection area is located at $s=-s_{\mathrm{ref}}$ consistently, ensuring that the localization on the incoming lanes is independent of the maneuver to be driven.
Moreover, this normalization makes the scene representation robust to slight changes in intersection geometry.

The second input feature $v$ denotes the scalar velocity of the corresponding vehicle normalized over the speed limit of the lane it is currently driving on.
To allow the network to sense immediate proximity of other vehicles, the input features are complemented by a distance measure $d$ based on the Mahalanobis distance \cite{de_maesschalck2000mahalanobis}.
The distance measured from vehicle~$i$ to vehicle~$j$ is calculated as
\begin{equation}
\label{eq:dist}
d_{ij} = \sqrt{(\boldsymbol{p}_j - \boldsymbol{p}_i)^T \boldsymbol{\varSigma}_i^{-1} (\boldsymbol{p}_j - \boldsymbol{p}_i)},
\end{equation}
where $\boldsymbol{p}_i$ denotes the position of vehicle $i$ in cartesian coordinates. The covariance matrix is given as
\begin{equation}
\boldsymbol{\varSigma}_i =
	\boldsymbol{R}_{\psi_i}
	\begin{bmatrix}
		l^2/4 & 0 \\
		0 & w^2/4
	\end{bmatrix}
	\boldsymbol{R}_{\psi_i}^T,
\end{equation}
with $l=\SI{5}{\meter}$ and $w=\SI{2}{\meter}$ describing the standardized length and width of a vehicle.
$\boldsymbol{R}_{\psi_i}$ denotes the 2D rotation matrix using the heading angle $\psi_i$.
To determine the input feature for a particular vehicle, the distance to each other vehicle is computed according to \eqref{eq:dist}, and the inverse of the minimum distance value is passed to the network.
Using the inverse instead of the plain distance value proved to yield better model convergence.

\subsection{Network Architecture}
\label{ssec:networkdetails}

In the present work, the behavioral control of vehicles is performed by applying a commanded longitudinal acceleration within the range $[a_{\mathrm{min}},\,a_{\mathrm{max}}]$, requiring an RL algorithm suited for continuous control.
We propose to use the twin delayed deep deterministic policy gradient (TD3) algorithm \cite{fujimoto2018addressing}, an extension of the deep deterministic policy gradient (DDPG) \cite{lillicrap2016continuous}.
Both methods are actor-critic RL algorithms for actions in continuous space.
TD3 consists of two function approximators, namely the \emph{actor} and the \emph{critic}.
Based on a given state and action input, the critic network is trained to predict the discounted reward as $Q$ value estimates.
The actor gets the current environment state as an input and outputs an action to be performed in the particular time step, optimized by using the critic output as the loss.

\begin{figure}
	\centering
	\includegraphics[width=\linewidth]{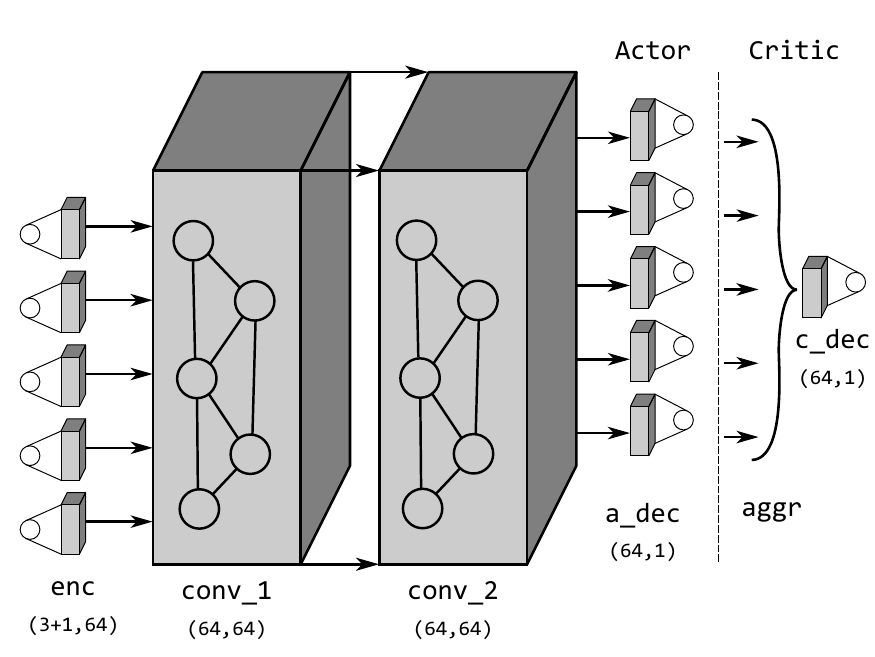}
	\caption{The graph neural network architecture is depicted, consisting of one dense encoder layer, two graph convolution layers, and one dense output layer. Below the layer identifiers, their dimensionalities are shown. The encoder has three input channels for the actor and four channels for the critic network.}
	\label{fig:network}
\end{figure}

The proposed graph neural network (GNN) architecture is depicted in Fig.~\ref{fig:network}.
For each vertex, the low-dimensional input features are first processed by a dense layer \texttt{enc}.
Note that this operation is performed individually for each vertex using shared parameters, disregarding the graph structure defined by the edges.
The encoded vertex features are then propagated through two relational graph convolution layers, \texttt{conv\_1} and \texttt{conv\_2}.
In contrast to simple graph convolution layers, multiple weight matrices corresponding to the different edge types are used for message passing \cite{gangemi2018modeling}.
During a forward pass, the hidden features of the vertices are propagated along the outgoing edges, while being multiplied by the respective weight matrix.
Hence, each node receives a variable amount of such messages that have to be integrated into its own feature vector.
This is done using a permutation invariant aggregation function like the element-wise maximum, mean, or sum.
In the present work, the maximum operation delivered the best results.
The update for the hidden feature vector of node $i$ is thus given as
\begin{equation}
\boldsymbol{h}_i^{(l+1)} = \sigma \left( \Sigma_{r \in U} \max_{j \in \mathcal{N}_i^r} \boldsymbol{W}_r^{(l)} \boldsymbol{h}_j^{(l)} + \boldsymbol{W}_0^{(l)} \boldsymbol{h}_i^{(l)} \right),
\end{equation}
where the set of neighbor nodes connected to the target node $i$ by incoming edges is denoted by $\mathcal{N}_i^r$.
The weight matrices for each edge type $r \in U$ are called $\boldsymbol{W}_r$, while the previous target node vector is multiplied by $\boldsymbol{W}_0$.
In case a vertex has no incoming edges (like $v_4$ or $v_7$ in Fig.~\ref{fig:scene_graph}), the node update coincides with a single dense layer on the node's own state.
Finally, the resulting feature vector is passed through a non-linear activation function, empirically chosen as rectified linear unit (ReLU).

While the actor network and the critic network are constructed analogously up to the point described above, they differ in the output layer, depicted on the right side of Fig.~\ref{fig:network}.
The actor network is responsible for deriving an action for each entity in the scene to be executed for a given time horizon, while the critic is in charge of estimating the Q~value for the entire graph.
With each vehicle being represented by a vertex in the graph, there is one regression target per vertex that describes the commanded acceleration for the corresponding vehicle in the actor network.
The latent feature output by the GNN is reduced to a single unit using a final dense layer \texttt{a\_dec}, whose weights are shared across nodes.
To limit the action output to a defined range, a tangens~hyperbolicus activation function is used on the output layer.
The normalized value range is subsequently mapped to an acceleration between \SI{-5}{\meter\per\second\squared} and \SI{+5}{\meter\per\second\squared}.
With the critic network's output being a performance measure in form of a single Q~value estimate, an aggregation function for the latent feature vector of all nodes is required.
The Q~values' range is not limited, hence a final dense layer \texttt{a\_dec} with linear activation is used as the output layer.
Because the critic network requires the chosen action in addition to the state representation, the action values are concatenated with all vertex input features.

\subsection{Reward Engineering}
\label{ssec:reward}

Apart from the network architecture described above, the RL algorithm requires a reward scheme to learn a reasonable behavior within the simulation environment.
The reward signal is composed of a weighted sum of reward components
\begin{equation}
R = \sum\limits_{k \in \mathcal{R}} w_k R_k,
\end{equation}
where the set of reward components is given as $\mathcal{R} = \{\mathrm{velocity,\,action,\,idle,\,proximity,\,collision}\}$.
The velocity reward is the main driver for learning a non-trivial solution through rewarding large velocities and is defined as
\begin{equation}
	R_\mathrm{velocity} = \begin{cases}
	1.25 \frac{v}{v_\mathrm{lim}}      &       \frac{v}{v_\mathrm{lim}} \le 0.8 \\
	1.0                                & 0.8 < \frac{v}{v_\mathrm{lim}} \le 1.0 \\
	6.0 - 5.0 \frac{v}{v_\mathrm{lim}} & 1.0 < \frac{v}{v_\mathrm{lim}},
	\end{cases}
\end{equation}
where $v_\mathrm{lim}$ describes the vehicle's lane speed limit.
Regularizing the model against applying large acceleration magnitudes is done by the action penalty that is defined as the negative absolute commanded acceleration.
When striving to avoid collisions, the simplest solution is to stop the whole traffic, which is not desirable.
Therefore, the idle penalty is set to $R_\mathrm{idle}=1$ in case all vehicles are standing still.
To teach the model to keep suitable safety distances to nearby vehicles, the proximity component is used to penalize actions that cause two vehicles to get dangerously close.
This penalty is calculated based on the aforementioned modified Mahalanobis distance measure (cf.~\eqref{eq:dist}), which takes the relative direction of the obstacle into account.
In the case that two vehicles collide, the collision penalty $R_\mathrm{collision}=1$ is used to let the model implicitly learn collision avoidance, while aborting the episode on the spot preventing further positive rewards from being accumulated.

\section{Experiments}
\label{sec:eval}

Training and evaluation of RL algorithms require a suited simulation environment.
For the task of behavioral planning in automated driving, the simulator should at least provide a kinematic vehicle model and reasonable interaction between vehicles.
In the present study, the open-source environment Highway-env \cite{highway-env} is used and slightly adapted to be employed for centralized multi-agent planning.
The simulation of vehicle kinematics is done according to the kinematic bicycle model \cite{kong2015kinematic}, which suffices for behavioral planning.
The graph-based scene representation and graph neural network layers are based on the PyTorch Geometric API \cite{pyg}.

\begin{table}
	\caption{Reward weights}
	\label{tab:rewardweights}
	\centering
	\begin{tabular}{lrrrrr}
		\toprule
		\textbf{Reward} & velocity & action & idle & proximity & collision \\ \midrule
		\textbf{Weight} &     0.03 &   0.01 & 0.01 &       0.2 &       1.0 \\ \bottomrule
	\end{tabular}
\end{table}

The choice of reward weights is based on a grid search that was conducted on a reduced variant of the simulation environment resembling the key behavior, while being much less computationally demanding.
The reward weights used throughout this study are given in Table~\ref{tab:rewardweights}.
During training, the latest model is evaluated on a separate validation environment for ten episodes every 5000~time steps.
The validation environment is constructed the same way as the training environment but initialized with a different seed.
Each time the validation shows a new best validation reward, the current model parameters are saved to disk.

We benchmark our approach against two baselines in synthetic simulation: static priority rules (PR) and a first-in-first-out scheme (FIFO), resembling the currently prevalent approach in real-world and a seminal AIM scheme.
Traffic obeying to static priority rules is simulated using the driver models provided by Highway-env, which rely on the intelligent driver model (IDM) \cite{treiber2000congested} for longitudinal control and additional logic for handling intersections.
We applied minor tweaks to the driver models to obtain reasonable results for more dense traffic:
\begin{itemize}
	\item The derivation of the commanded acceleration is modified to correctly handle a target speed of zero.
	\item Scheduling at an intersection is based on vehicle priorities that are inferred from their intended maneuver instead of the current lane priority.
	\item The yielding logic is extended to respect a specific stop point in front of the intersection.
\end{itemize}
The FIFO scheme, on the other hand, prioritizes the incoming vehicles based on their distance to the intersection.
Thereby, non-conflicting paths can be used at the same time, possibly allowing multiple vehicles on the intersection at a given time.
Note that this policy does not enforce a strict FIFO ordering on the whole intersection, but rather groups of conflicting paths, which leads to a considerable performance increase.

The way of generating simulated traffic differs between training and evaluation runs.
During training, vehicles are being spawned on all incoming lanes featuring enough space at a certain probability with their destination also being chosen randomly.
In the course of training, the spawn probability is continuously increased until it saturates at \SI{5}{\percent} per time step.
Thereby, the intersection is kept busy and allows the RL algorithm to obtain meaningful data samples to learn from.
The evaluation runs are based on scenario definitions that are generated using a slightly different scheme.
In that case, the time between vehicles appearing on a particular lane is governed by a shifted exponential distribution.
This resembles a Poisson process, where the shift on the distribution of the spawn period ensures a minimum distance between vehicles.
If a traffic jam has formed so that there is no space for a vehicle to be spawned, the generation is suspended to prevent immediate collisions.
For each scenario to be generated, the desired vehicle rate is chosen randomly from a uniform distribution over the interval $[0.2,\,0.4]$~vehicles per second and major road lane.
The vehicle rate on the minor road lanes is set to half of that value.
In both training and evaluation, the initial vehicle velocity is chosen uniformly between \SI{60}{\percent} and \SI{100}{\percent} of the corresponding lane speed limit.

\subsection{Synthetic Simulation}
\label{ssec:synthetic}

\begin{figure}
	\centering
	\includegraphics[width=\linewidth]{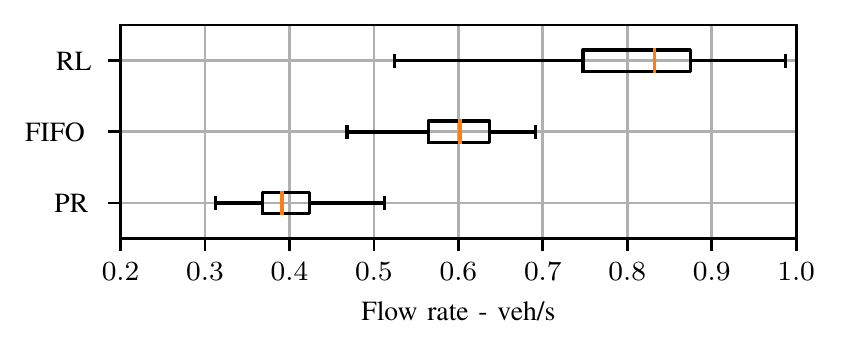}
	\caption{Flow rate during various evaluation runs by using static priority rules (PR), the FIFO~policy, and the RL~planner.}
	\label{fig:syn_flow_rate}
\end{figure}

\begin{figure}
	\centering
	\includegraphics[width=\linewidth]{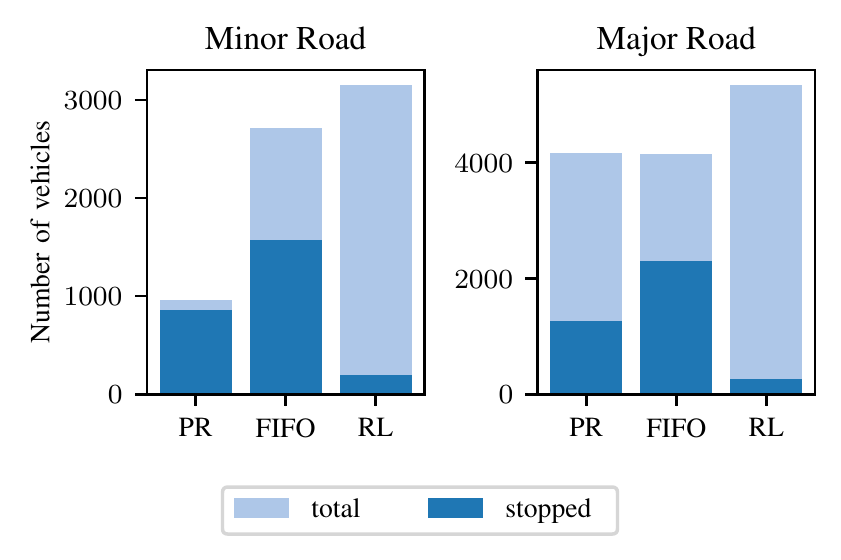}
	\caption{The number of vehicles approaching the intersection from the major and minor road as well as the ratio of those that had to stop when governed by static priority rules (PR), the FIFO~policy, and the RL~planner.}
	\label{fig:syn_road_stops}
\end{figure}

All experiments within this subsection were performed on a four-way intersection, whose layout is analogue to the one depicted in Fig.~\ref{fig:scene_graph}.
The \emph{flow rate} describes the number of vehicles that cross an intersection (or other road infrastructure) during a given time frame.
Figure~\ref{fig:syn_flow_rate} shows the flow rate distribution over 100 evaluation runs (each of \SI{100}{\second} length) of varying traffic density.
It can be observed that already the rather simple FIFO~scheme achieves a benefit over static priority rules, while the learned RL~planner outperforms both baselines regarding the median values.
To further investigate the performance gain, we analyze the ratio of vehicles that had to stop during the maneuver.
A vehicle trajectory is considered to contain a stop, if the velocity falls below \SI{0.3}{\meter\per\second} for at least one time step.
This threshold was chosen due to numeric reasons.
By categorizing the vehicles by their incoming road priority, the effect on traffic approaching from a minor road becomes apparent, as depicted in Fig.~\ref{fig:syn_road_stops}.
Clearly, the static priority rules induce a significant traffic buildup on the minor road that forces nearly all vehicles to stop.
The FIFO~policy manages to let more vehicles from the minor road pass the intersection, but in turn causes a large proportion of stops also on the major road.
In contrast, the RL~planner succeeds to get a large amount of vehicles across the intersection, while keeping the traffic flow mostly intact.
This behavior might be explained by the planner's learned ability to adapt the vehicles' velocities early to fit into an emergent gap.
It should be mentioned that the RL~planner cannot completely eliminate collisions, as denoted in the first row of Table~\ref{tab:collisions}.
However, the occurrence is extremely rare, making it very challenging to further reduce them, given the RL~algorithm has to implicitly learn it via the reward signal.
In practice, the remaining failure cases are not an issue, because the cooperative maneuver will only be advertised to the connected vehicles, if it fulfills sanity checks like being collision-free.
In the case no viable cooperative plan was found, the vehicles simply resort to local planning.

\begin{table}
	\caption{Collision rates}
	\label{tab:collisions}
	\centering
	\begin{tabular}{lrrr}
		\toprule
		\textbf{Intersection} & \textbf{Priority rules} & \textbf{FIFO scheme} &  \textbf{RL planner} \\ \toprule
		Synthetic 4-way       &      \SI{0.0}{\percent} &   \SI{0.0}{\percent} & \SI{0.028}{\percent} \\ \midrule
		inD                   &      \SI{0.0}{\percent} & \SI{1.918}{\percent} & \SI{0.584}{\percent} \\ \bottomrule
	\end{tabular}
\end{table}

\subsection{Simulation Based on Real-World Traffic Data}
\label{ssec:realworld}

\begin{figure}
	\centering
	\includegraphics[width=\linewidth]{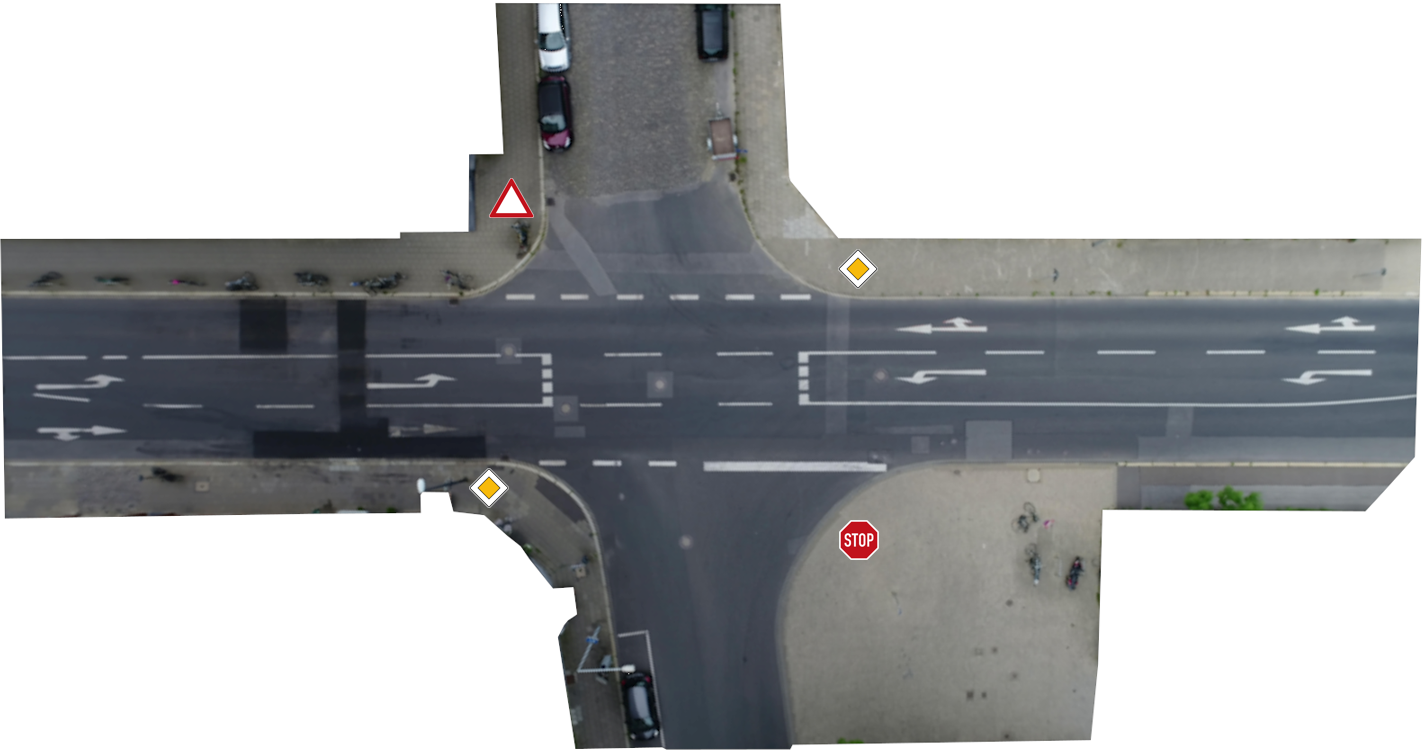}
	\caption{Bird's-eye view of the intersection, where the real-world vehicle tracks taken from the inD dataset were collected (image adopted from \cite{bock2020ind}).}
	\label{fig:ind_intersection}
\end{figure}

Apart from the simulation based on synthetic data, we also evaluated the cooperative planning scheme on real-world urban traffic data taken from the inD dataset \cite{bock2020ind}.
The dataset contains tracks of vehicles and vulnerable road users that were recorded at four urban intersections in Germany.
We selected a four-way intersection connecting a priority road with a minor road that is managed by static priority rules.
The major road also features isolated lanes for turning left, as depicted in Fig.~\ref{fig:ind_intersection}.
Simulating the traffic according to the cooperative planning approach on this intersection makes the following assumptions inevitable.
Firstly, the intersection geometry is only approximated in simulation.
However, this is not an issue for behavioral planning, which is mostly independent of road geometry.
Considering the road curvature, vehicles may not traverse it with arbitrary speed, which is ensured by defining lane-dependent velocity limits.
As the dynamic properties of the vehicles shown in the dataset are unknown, a default parameter set is used in simulation.
Moreover, the real-world tracks deviate from the lane center lines that are used for guiding the simulated vehicles.
The recorded intersection is also used by vulnerable road users like pedestrians and bicyclists that cannot be modeled in the simulation as of now.

Compared to the synthetic simulation, not all metrics are viable for evaluation when using the dataset as the baseline.
The flow rate, for instance, cannot be improved by any intersection management system, because the number of vehicles in the scene is specified by the dataset.
In total, the used dataset excerpt provides 2446~vehicle tracks over a recording time of 3.08~hours, resulting in an average traffic density of \SI{794}{\vehicle\per\hour} or \SI{0.221}{\vehicle\per\second}.
Compared to the flow rates obtained in the synthetic simulations, those numbers are rather small, which might indicate that there are not many interesting situations during most of the recording.
Evaluating the RL~planner and FIFO~policy on real-world data is performed by spawning vehicles in simulation according to the appearance time in the dataset and subsequently simulating their motion based on the vehicle models.

\begin{figure}
	\centering
	\includegraphics[width=\linewidth]{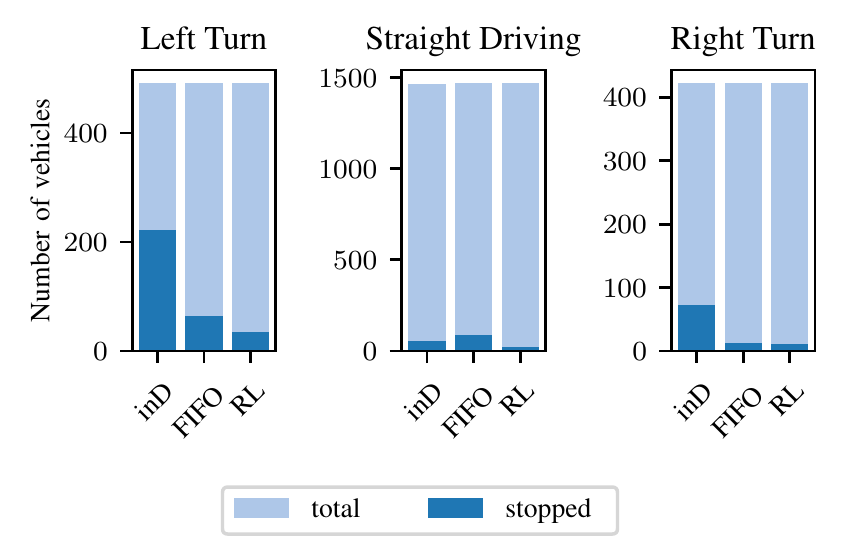}
	\caption{The number of vehicles that perform a certain maneuver and ratio of those that had to stop in real-world data (inD), using the FIFO~policy, and the RL~planner.}
	\label{fig:ind_turn_stops}
\end{figure}

Figure~\ref{fig:ind_turn_stops} compares the number of vehicles that had to come to a complete stop categorized into the maneuvers left turn, straight driving, and right turn.
The total number of vehicles being managed by the FIFO~policy and the RL~planner has to be identical to the amount given by the dataset recording.
It is clearly visible from the real-world data that many left turning vehicles have to come to a stop before being able to safely pass the intersection.
Note that although one minor road access is governed by a stop sign, the real-world recordings show that by far not all road users obey to it.
The FIFO scheme naturally distributes the stops to all maneuvers including vehicles driving straight on.
Meanwhile, the RL~planner is able to avoid the vast majority of stops and maintains a smooth flow of traffic.

As it can be seen in the last row of Table~\ref{tab:collisions}, both the FIFO and the RL~planner suffer from an increased collision rate.
This can, at least in parts, be attributed to the way the real-world tracks are mapped to simulation.
Especially the minor road is only depicted for a very short distance in front of the intersection (cf. Fig.~\ref{fig:ind_intersection}), which makes it difficult for the planner to influence the incoming vehicles before they enter the intersection area.
In case one vehicle waits at the stop point while a second vehicle enters the scene at high speed, a collision might be inevitable.
This is merely an issue of the evaluation and does not diminish the remarkable improvement in traffic efficiency that is raised by the RL~planner.

\section{Conclusion}
\label{sec:conclusion}

In this work, a novel multi-agent behavioral planning scheme for connected automated vehicles at urban intersections was presented.
We chose a reinforcement learning algorithm to leverage recent advances in machine learning while evading the need for ground truth data that is virtually unavailable for cooperative maneuvers.
The developed graph-based input representation effectively encodes the semantic environment at the operational area.
By employing graph neural networks, our approach confidently handles the varying number of vehicles in the scene.
The proposed approach was evaluated in synthetic simulation and additionally based on real-world traffic data.
Compared to static priority rules and a FIFO~scheme as baselines, the learned planner increases the vehicle throughput significantly.
In addition, the number of induced stops is reduced which indicates better traffic flow.

The proposed behavioral planning framework can serve as a sound foundation for solving more sophisticated planning problems.
In the future, we plan to extend this work to be applicable to intersection layouts that were not seen during training.
Moreover, cooperative planning in mixed traffic, i.e. human drivers and automated vehicles sharing the road, shall be addressed.

\balance
\bibliographystyle{IEEEtran}
\bibliography{references}

\end{document}